\documentclass[letterpaper, 10pt, conference]{ieeeconf}

\IEEEoverridecommandlockouts   
\usepackage{array}
\usepackage{amsmath}
\usepackage{graphicx}
\usepackage{caption}
\usepackage{siunitx}
\usepackage{booktabs}
\usepackage{multirow}
\usepackage{xspace}
\let\labelindent\relax  
\usepackage{enumitem}
\usepackage{wrapfig}
\usepackage{placeins}
\usepackage{subcaption}
\usepackage{float}
\usepackage{stfloats}   
\usepackage{needspace}
\usepackage[most]{tcolorbox}
\usepackage{microtype}

\usepackage{needspace}
\usepackage{etoolbox}
\usepackage[T1]{fontenc}
\usepackage{fancyvrb}
\usepackage{fvextra}   

\newcommand{\redboxfont}{\fontsize{8pt}{9pt}\selectfont}
\fvset{fontsize=\redboxfont,formatcom=\normalfont}
\usepackage{url}
\usepackage{xurl}
\usepackage[hidelinks]{hyperref}

\apptocmd{\thebibliography}{%
  \setlength{\itemsep}{0pt}%
  \setlength{\parsep}{0pt}%
  \setlength{\parskip}{0pt}%
  \interlinepenalty=0\relax
}{}{\PackageError{InSight}{Could not set bibliography spacing}{}}

\setlist[itemize]{topsep=1pt, itemsep=1pt, parsep=0pt, partopsep=0pt, leftmargin=*}
\setlist[enumerate]{topsep=1pt, itemsep=1pt, parsep=0pt, partopsep=0pt, leftmargin=*}

\newcommand{\method}{\textsc{InSight}\xspace}
\newcommand{\rwheading}[1]{\par\noindent\textbf{#1.}\enspace}

\newif\ifincludeappendix
\includeappendixtrue

\definecolor{StanfordCardinal}{HTML}{9F1C1F}
\definecolor{StanfordLightRed}{HTML}{F4E6E7}
\definecolor{StanfordMidRed}{HTML}{D8A6A7}

\tcbset{
  stanfordstyle/.style={
    colback=StanfordLightRed,
    colframe=StanfordCardinal,
    colbacktitle=StanfordMidRed,
    coltitle=black,
    boxrule=1pt, arc=3pt,
    left=6pt, right=6pt, top=5pt, bottom=5pt,
    enhanced,
    fonttitle=\redboxfont\bfseries,
    fontupper=\redboxfont,
  },
}

\newcommand{\knownbadge}{%
  \tcbox[
    on line, colback=white, colframe=StanfordCardinal,
    boxrule=0.4pt, arc=2pt,
    left=2pt, right=2pt, top=0pt, bottom=0pt
  ]{\redboxfont KNOWN}%
}

\title{\fontsize{20pt}{25pt}\selectfont
InSight: Self-Guided Skill Acquisition via Steerable VLAs}

\author{Maggie Wang$^{1}$, Lars Osterberg$^{1}$, Stephen Tian$^{1}$,
        Ola Shorinwa$^{2}$,\\[2pt]
        Joseph Bowkett$^{3}$, Daniel Pastor$^{3}$, Jiajun Wu$^{1}$, and Mac Schwager$^{1}$%
\thanks{$^{1}$Stanford University.}%
\thanks{$^{2}$Princeton University.}%
\thanks{$^{3}$NASA Jet Propulsion Laboratory.}%
}

\begin{document}


\IEEEaftertitletext{%
\vspace{-0.5em}
\centering
\includegraphics[width=\textwidth]{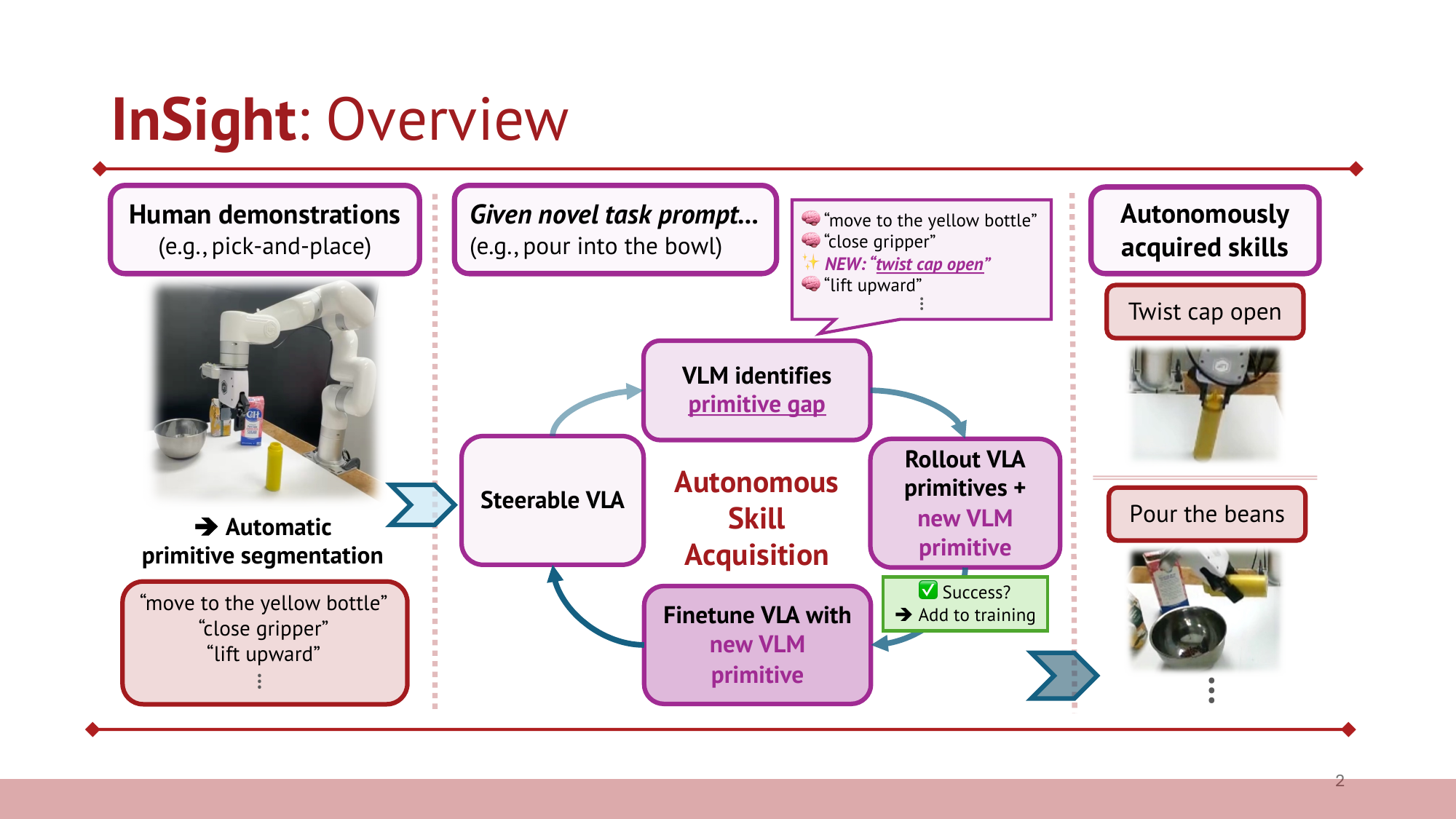}
\vspace{-0.5em}
\captionof{figure}{\small{\textbf{Overview of \method.} 
(Left)~Human demonstrations are automatically segmented into primitive-labeled trajectories to fine-tune a VLA to be steerable via these primitive labels. 
(Middle)~Given a new task, \method uses a VLM to create a plan consisting of \textit{known primitives}, executed by the primitive-steerable VLA, and \textit{new primitives}, executed through VLM-parameterized low-level control. \method autonomously collects task rollouts and retrains the VLA on the original data with new-primitive segments from successful rollouts, expanding the VLA's primitive vocabulary. 
(Right)~The newly acquired primitives (e.g., twisting, pouring) can be composed to perform new tasks without additional human demonstrations.}}
\label{fig:overview}
}

\maketitle

\setlength{\textfloatsep}{3pt}   
\setlength{\intextsep}{3pt}      
\setlength{\floatsep}{3pt}       
\setlength{\belowcaptionskip}{0pt}

\thispagestyle{empty}
\pagestyle{empty}


\begin{abstract}
    Vision-language-action (VLA) models excel at robot manipulation via imitation learning, but adapting them to new tasks often requires additional human demonstrations, which can be costly or infeasible.
    Meanwhile, vision-language models (VLMs) offer semantic task understanding but lack the physical grounding required for execution. 
    To bridge this gap, we present \method, a framework for self-guided skill acquisition that uses a VLM to identify primitives missing from a VLA's repertoire, grounds the VLM's proposals through robot execution, and distills new primitives from successful rollouts into the VLA.
    Primitive steerability---the ability to execute and terminate primitives on command---enables the robot to reuse known primitives while collecting training data for missing primitives without requiring full-task human demonstrations for each new task.
    \method has two stages:
    (1) a VLM automatically segments existing demonstrations into primitive-labeled trajectories to fine-tune a primitive-steerable VLA,
    and
    (2) the VLM plans a sequence of known primitives executed by the VLA and new primitives attempted by VLM-parameterized low-level controllers.
    New-primitive segments from successful task rollouts are added to the training data, and the VLA is retrained.
    The adapted VLA can then reliably execute new skills using the acquired primitives, without per-primitive VLM calls.
    We evaluate \method on six simulated and real-world tasks with no human demonstrations of target skills, including block flipping, drawer closing, sweeping, twisting, and pouring.
    On hardware, acquired twisting and pouring skills achieve $92\%$ and $96\%$ success, versus $32\%$ and $16\%$ for a zero-shot CaP-X baseline.
    Composing both skills into a 14-primitive task achieves $80\%$ success with no combined-task demonstrations. \textbf{Project website:} \url{https://insight-vla.github.io/}.

\end{abstract}




\section{Introduction}

Teaching robots new manipulation skills is expensive. 
Vision-language-action (VLA) models have made progress toward general-purpose manipulation~\cite{kim_openvla_2024, intelligence2025pi05visionlanguageactionmodelopenworld, nvidia2025gr00tn1openfoundation}, but adapting them to a new task still requires collecting human demonstrations, which often incurs substantial human effort, limiting scalability. 
Vision-language models (VLMs), meanwhile, have commonsense knowledge that can be used to propose how to perform a wide variety of manipulation tasks. However, using these proposals directly for robot control presents two challenges: the proposed motions may be physically infeasible or prone to failure during execution, and repeated VLM calls can introduce substantial test-time latency~\cite{liang_code_2023, fu_cap-x_2026}.

Consider a robot operating on the surface of Mars that has been trained to scoop rocks for sample collection. If a dust storm deposits debris onto its solar panels, the robot may need to clear the panels~\cite{noauthor_nasas_2022}, yet a VLA trained only to scoop may fail to execute the required sweeping behavior because no demonstrations of sweeping were provided. While a VLM can propose sweeping actions, the proposals may not be feasible on the real robot.
Acquiring new skills through interaction can also be difficult: simulation-based reinforcement learning (RL) often faces sim-to-real gaps, while real-world RL largely remains impractical due to sample complexity and safety constraints~\cite{kalashnikov2018qtoptscalabledeepreinforcement, wagenmaker_steering_2025}.

Our key insight is that manipulation skills are inherently compositional: new manipulation skills are rarely entirely new, but instead reuse previously seen primitives in new combinations~\cite{gu_learning_2026, garrett_integrated_2020}. 
For instance, sweeping and scooping share approach and lowering primitives, but differ in the lateral pushing primitive. 
Similarly, flipping a block reuses the same grasp-and-lift sequence from pick-and-place but adds a rotation primitive not present in those demonstrations. 
This compositional structure suggests that skill acquisition reduces to learning targeted missing primitives while harnessing existing primitives known by the VLA---without necessarily requiring full-task human demonstrations. 
To reuse known primitives while collecting training data for missing primitives, a VLA must be able to execute and terminate primitives on command---a capability referred to as \emph{primitive steerability}.
However, prevailing VLA architectures entangle atomic primitives within the full task instruction~\cite{chen_steerable_2026, smith_steer_2025}, making the VLA challenging to steer at the primitive level.


%


Efficiently acquiring a new skill requires not only executing known primitives (e.g., with a capable VLA), but also recognizing what primitives are missing to achieve a task, which a VLM can provide.
While recent work uses VLMs and LLMs as planners, coding agents, or trajectory-generation modules that compose pre-trained or pre-defined primitives at test time~\cite{liang_code_2023, ahn_as_2022, huang_voxposer_2023, fu_cap-x_2026}, these methods extend the robot's behavior through repetitive test-time reasoning without updating the learned policy, degrading inference-time efficiency. 
To address this limitation, we propose a different role for the VLM: not only as a test-time planner over existing skills, but as an \textit{active agent} for identifying \textit{missing} primitives, generating successful robot rollouts, and adding those rollouts back to the VLA by retraining to extend its skill capabilities.

This process is analogous to how humans encounter a new scenario: we understand what skills we can already perform, and thus recognize when current skills are insufficient. We then reason about what new capability would bridge the gap and learn using targeted practice. The acquired skill can then be stored as a reusable capability for future tasks, thus enabling continual, lifelong learning. 

To this end, we present \method, a framework for \textbf{skill acquisition via steerable VLAs}. Fig.~\ref{fig:overview} summarizes the overall pipeline, from primitive segmentation of human demonstrations to VLM-guided acquisition and modular composition of new primitives. We show how a VLA can be made steerable at the level of composable manipulation primitives and then autonomously extended when a new task requires a missing primitive. Our contributions are as follows: 
\begin{itemize}
    \item An \textbf{automatic primitive segmentation} pipeline that decomposes teleoperated demonstrations into labeled primitives without manual annotation, enabling primitive-level VLA steerability.
    \item A \textbf{VLM-guided primitive acquisition} loop that identifies the primitives a VLA is missing for a new task, grounds the VLM's proposals through robot execution, and retrains the VLA on autonomously generated demonstrations to accomplish new skills.
\end{itemize}

We validate \method across six tasks in simulation and on hardware, including block flipping, drawer closing, sweeping, twisting, and pouring. \method enables autonomous skill acquisition with zero target-skill human demonstrations, achieving up to 96\% success on tasks such as pouring, and 80\% success on a complex 14-primitive long-horizon task while retaining full performance on original base skills. 



\section{Related Work}

\begin{figure*}[b]
\centering
\centering
  \includegraphics[width=1\textwidth]{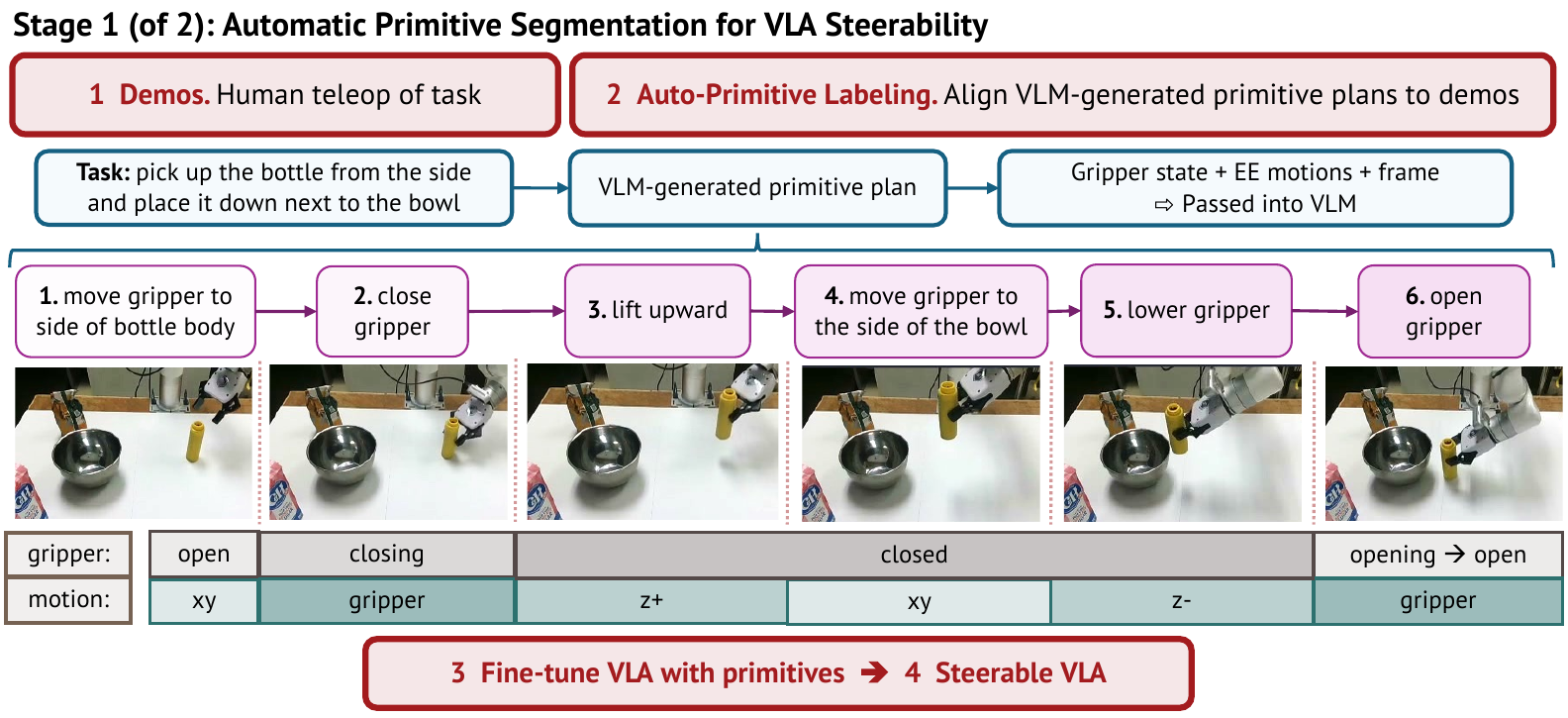}
  \caption{\small{\textbf{Stage 1: automatic primitive segmentation.} (1)~Human collects teleoperation data of a task. (2)~These demonstrations are decomposed into labeled primitives by aligning a VLM-generated primitive plan with gripper-state transitions and dominant end-effector motion (shown below the frames). (3)~The primitive-labeled trajectories are used as individual training episodes to fine-tune the VLA. (4)~The retrained VLA can execute and terminate individual primitives on command, providing the \textit{primitive-steerable} interface used for skill acquisition in Stage 2 (Fig.~\ref{fig:methods-stage2}). In this example, pick-and-place demonstrations are segmented into six labeled primitives.
  }}
  \label{fig:methods-stage1}
\end{figure*}

\label{sec:citations}

\rwheading{Vision-Language-Action Models and Steerable Policies}
VLAs map visual observations and language instructions to robot actions, learning end-to-end control from language-labeled robot data~\cite{kim_openvla_2024, intelligence2025pi05visionlanguageactionmodelopenworld}.
Recent work explores finer-grained language conditioning for more steerable control interfaces~\cite{intelligence2026pi07steerablegeneralistrobotic, chen_steerable_2026, liu_vls_2026}. STEER~\cite{smith_steer_2025} shows that dense language relabeling can expose an expressive low-level control interface, while Steerable Policies~\cite{chen_steerable_2026} similarly uses language-conditioned primitives to guide behavior at test time.
These methods establish steerability as a useful test-time control interface, but treat the resulting primitive set as fixed. \method instead uses steerability as the mechanism for expanding the policy's reusable skill set. After training a steerable VLA with a set of known primitives, \method identifies missing primitives for a new task, generates successful rollouts for those primitives, and adds them back into the VLA's training data. 

\rwheading{Skill Decomposition and Composition}
Hierarchical robot learning methods decompose manipulation into reusable skills that can be sequenced for long-horizon tasks~\cite{gutierrez2026movementprimitivesroboticscomprehensive, lee_equivariant_2023, liu_learning_2025, gu_learning_2026}. Bottom-up skill discovery methods extract reusable primitives from unsegmented demonstrations or reward-free interaction using clustering, representation learning, hierarchical imitation learning, or factorized skill spaces~\cite{zhu_bottom-up_2022, pmlr-v229-xu23a, cathomen_divide_2025, nie2026stack}. These methods discover primitives that already exist in the available data, whereas \method couples skill decomposition with autonomous acquisition to extract primitives from demonstrations and acquire new primitives with VLM guidance.

\rwheading{LLM/VLM-Guided and Agentic Robotics}
Foundation models have been used to provide high-level semantic guidance for robot execution~\cite{ahn_as_2022, huang_voxposer_2023}.
Code-as-Policies~\cite{liang_code_2023} and CaP-X~\cite{fu_cap-x_2026} compose programmatic skills through test-time reasoning without updating the policy, while 
Hi Robot \cite{shi_hi_2025} and BOSS \cite{zhang2023boss} use hierarchical or language-guided composition of existing capabilities.
Concurrent work such as ENPIRE~\cite{xiao2026enpireagenticrobotpolicy}, ASPIRE~\cite{lu2026aspireagenticskillsdiscovery}, and RATs~\cite{zhang2026playfulagenticrobotlearning} explores agentic systems that persistently expand or improve robot capabilities through self-directed skill acquisition, program refinement, or automated policy optimization.
\method instead uses a VLM to expand the skill repertoire of pretrained VLAs by identifying missing primitives, generating and evaluating targeted robot rollouts; successful new primitives are then distilled into the policy.

\rwheading{Autonomous Skill Acquisition}
RL studies how agents can acquire skills through interaction. In real-world manipulation, RL typically demands dense rewards, substantial real-world interaction data, and a narrow sim-to-real gap~\cite{xu_rldg_2024}, which are generally challenging to achieve. Recent work reduces human supervision through foundation model-generated rewards~\cite{zhang_rewind_2025, ma_eureka_2024, zhao_agentic_2024}, automated demonstration data~\cite{mandlekar_mimicgen_2023, duan_manipulate-anything_2024}, or language-labeled trajectories~\cite{ha_scaling_2023} for robot policy learning. PLD~\cite{xiao2026selfimproving} uses residual RL to generate successful trajectories from states visited by a base VLA and distills them into the VLA for policy improvement.
\method instead identifies missing primitives, rolls out the full task using the steerable VLA on known primitives and VLM-parameterized control on the gaps, and distills only the successful new-primitive segments into the VLA for compositional reuse in new tasks.

\rwheading{Continual, Lifelong Skill Learning}
A related line of work studies how robots can continually acquire new skills while retaining prior skills~\cite{cheng_continual_2025}. Recent methods address this through structured knowledge reuse, including a continually evolving knowledge space with expert routing~\cite{wu_continually_2025} and semantic skill subspaces~\cite{wang_lifelong_2026}.
\method takes a data-centric approach by retraining a single VLA jointly on the original and newly acquired primitives, so existing primitives remain supervised as the vocabulary grows.


\section{Skill Acquisition via Steerable VLAs}
\label{sec:methods}

\method operates in two stages: (1) training a steerable VLA on automatically segmented primitives, and (2) autonomous skill acquisition, where a VLM orchestrates the full loop of primitive gap identification, data generation, success evaluation, and retraining. Fig.~\ref{fig:methods-stage1} shows the automatic primitive segmentation stage, and Fig.~\ref{fig:methods-stage2} shows the VLM-guided skill acquisition loop.






\subsection{Definitions}

In this paper, we define a \textbf{skill} as a target capability described by a language instruction (e.g., \textit{twist open the bottle and pour its contents into the bowl}). A \textbf{plan} is the sequence of primitives that the VLM planner generates to complete a skill. 

A \textbf{primitive} is a reusable action segment that the VLA produces when conditioned on its language label. 
Following the task and motion planning (TAMP) terminology~\cite{garrett_integrated_2020}, each primitive has an intended \textit{effect} on the world state described by its language label. \method does not explicitly represent preconditions; instead, the VLA's ability to execute a primitive from a given state is bounded by its training distribution, and the VLM planner determines the order in which primitives are invoked.
In our setting, primitives are designed to have a single dominant motion mode (e.g., translation or rotation along an axis or a gripper transition). Each primitive is associated with a natural-language label that describes its semantic effect (\textit{move gripper to the yellow bottle}, \textit{lift upward}, \textit{twist}, \textit{pour}). The VLA executes each primitive end-to-end until a termination signal is triggered (e.g., a learned progress detector crosses a completion threshold).

Let $\mathcal{V}$ be the policy's primitive vocabulary, or the set of primitive labels for which the VLA has been trained. 
We call a VLA \textbf{primitive-steerable} if, given a primitive label $p \in \mathcal{V}$ and an initial state in $p$'s distribution, it executes $p$ and terminates when $p$'s intended effect is achieved, independent of the full-task instruction that $p$ was originally demonstrated under (Sec.~\ref{sec:steerability}). In our work, steerability is a property of the control interface (not of task performance). Steerability enables the execution of a task at the primitive level without new full-task demonstrations, which is foundational to the acquisition loop in Stage 2.

Given a plan $\mathcal{P} = (p_1, \dots, p_n)$ generated by the VLM planner for a skill $s$, a \textbf{primitive gap} is any $p_i \in \mathcal{P}$ such that $p_i \notin \mathcal{V}$. 
\method autonomously acquires successful rollouts for each primitive gap. After these rollouts are added to the training set and the VLA is retrained, $\mathcal{V}$ expands and future plans can invoke the acquired primitives as known capabilities rather than primitive gaps. 
This framework enables new skills to be realized without additional human demonstrations. We define a \textbf{new} task operationally as one the fine-tuned VLA cannot perform in the environment without further training, regardless of whether similar tasks appear in the pretraining data; for example, VLAs fine-tuned with segmented pick-and-place demonstrations achieve $0/5$ twist successes before acquisition.

\subsection{Stage 1: Training a Primitive-Steerable VLA}
\label{sec:methods-stage1}

As shown in Fig.~\ref{fig:methods-stage1}, we automatically segment teleoperated demonstrations into labeled primitives without manual annotation.

Given a task description, a VLM produces an ordered sequence of expected primitives. We give the VLM a set of example primitives (e.g., ``close gripper'', ``lift upward'') to provide examples of the granularity of primitives. The primitive set is extended whenever the VLM proposes a new primitive needed to complete the task.

\begin{figure*}[t]
\centering
\centering
  \includegraphics[width=1.0\textwidth]{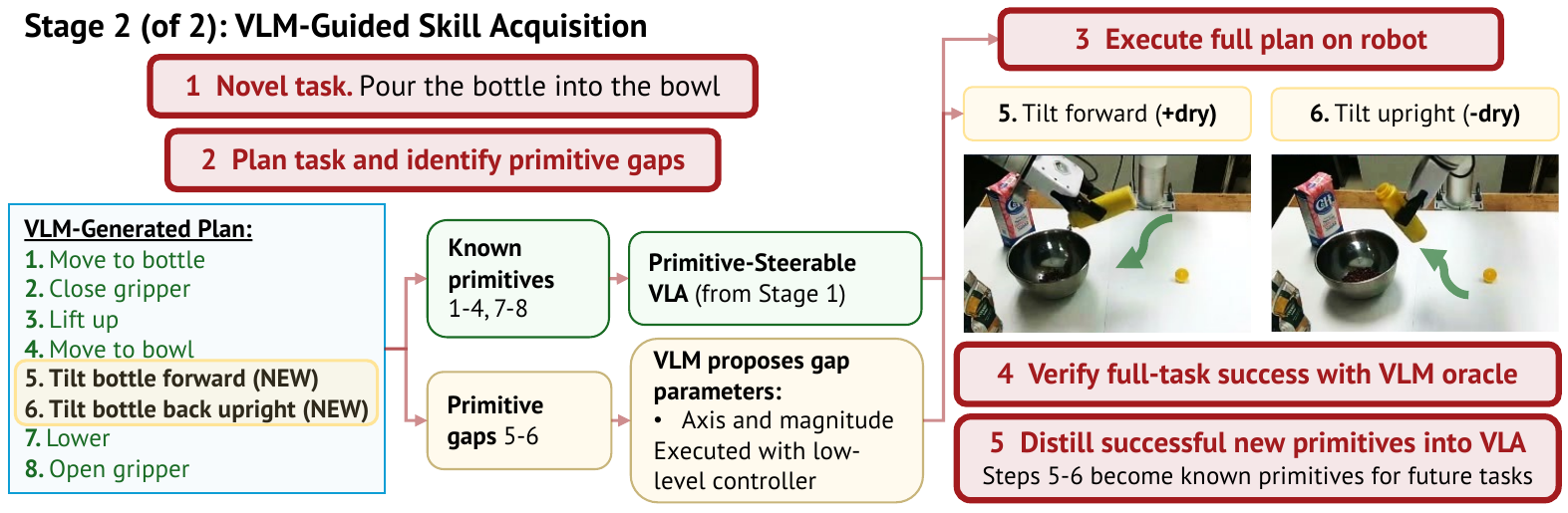}
  \caption{\small{\textbf{Stage 2: VLM-guided skill acquisition.} (1)~Given a new task, (2)~the VLM builds a plan and flags any primitive missing from the VLA's current vocabulary as a \textit{primitive gap}. (3)~During a rollout, known primitives are executed by the steerable VLA, while each primitive gap is executed by a low-level controller parameterized by a VLM-proposed motion axis and signed magnitude. (4)~A VLM oracle verifies task success, and new-primitive segments from successful rollouts are added to the training set and distilled into the VLA, becoming known primitives for future tasks. In this example, \method acquires the \textit{tilt bottle forward} and \textit{tilt bottle back upright} primitives needed to pour the bottle into the bowl.}}
  \label{fig:methods-stage2}
\end{figure*}

We segment boundaries at gripper open and close transitions from gripper command velocities. The end-effector pose, derived motion magnitudes (in $xy$ and $z$), and a dominant-axis tag (one of $\{xy, z, rxy, rz\}$) are passed to the VLM to match each frame with its associated primitive name.

The segmented primitives are used to fine-tune a pretrained VLA using LoRA \cite{hu2021loralowrankadaptationlarge}\ifincludeappendix\ (additional details in Appendix~\ref{app:implementation_details})\fi.
Each primitive segment is a separate training episode conditioned on its primitive label as the language prompt. To provide a primitive-level termination signal, we add a learned progress channel to the action space, with progress labels in $[0, 1)$ within each primitive.

\subsection{Stage 2: VLM-Guided Skill Acquisition}

Given a steerable VLA trained on a base set of primitives, \method autonomously expands the skill set when presented with a new task that requires missing primitives. A flowchart of this system is shown in Fig.~\ref{fig:methods-stage2}.

First, the VLM decomposes the task into a primitive sequence and compares against the known primitive vocabulary. Primitives not in the vocabulary are flagged as \textbf{primitive gaps}. Each gap is constrained as a single-axis motion, so tasks requiring multiple distinct motions (e.g., tilt forward and then tilt back) produce multiple primitive gaps rather than a single composite primitive.

Next, each primitive gap is parameterized by an axis (translation $\{dx, dy, dz\}$ or rotation $\{drx, dry, drz\}$) and a signed magnitude, which the low-level controller executes with a fixed speed. A pre-execution VLM call analyzes the current scene and proposes the parameters. For chained gaps within a single rollout, the prior gap's parameters are passed to the VLM to ensure a consistent axis and magnitude for paired motions (e.g., pour-forward then tilt-back-upright).

To acquire a new primitive, each plan execution produces a sequence of (known and/or new) primitives, which is rolled out by the robot. A post-plan VLM oracle compares the scene images before and after skill $s$ is run to judge full task success. Successful new primitives within $s$ are added to the VLA training dataset. After $n$ rollouts of the skill, we combine the existing primitive vocabulary $\mathcal{V}$ with the new primitive(s) $p$ to retrain the VLA. The retrained model can execute the full composed task autonomously, with the VLM acting only as a high-level planner to sequence the primitives. 

\section{Experiments and Analyses}

We evaluate \method across simulation and real-world manipulation tasks designed to test whether (1) missing primitives can be acquired without target-skill demonstrations, (2) acquired primitives are retained and can be composed, and (3) distilling acquired primitives into the VLA improves over executing them directly with the VLM at test time. Simulation success is measured from ground-truth state. Real-world success is judged by a human against binary task criteria, while the VLM oracle is used to filter acquisition rollouts for training.

In simulation, we use a 7DoF Franka Panda in the LIBERO~\cite{liu_libero_2023} environment to study block flipping from pick-and-place demonstrations to measure how performance improves as autonomously acquired rotate-block primitives are added, and drawer closing from drawer-opening demonstrations to test whether \method can acquire a missing primitive from an initial state absent from its fine-tuning demonstrations.

On hardware, we use a 6DoF UFactory xArm to evaluate bottle twisting and pouring to compare against a Code-as-Policies-style zero-shot baseline~\cite{fu_cap-x_2026}, and then compose the individually acquired twist and pour primitives along with the base pick-and-place skills into a long-horizon twist-then-pour task. We measure whether the unified VLA policy retains its original pick-and-place skills after new primitives are added. Finally, we evaluate whether \method extends to contact-rich, non-prehensile motions by acquiring a sweeping primitive from scooping demonstrations.

\subsection{Simulation: Block Flipping from \textit{Only} Pick-and-Place Human Demos}
\label{sec:block_flip}

\begin{figure}[t]
\centering
\includegraphics[width=\linewidth]{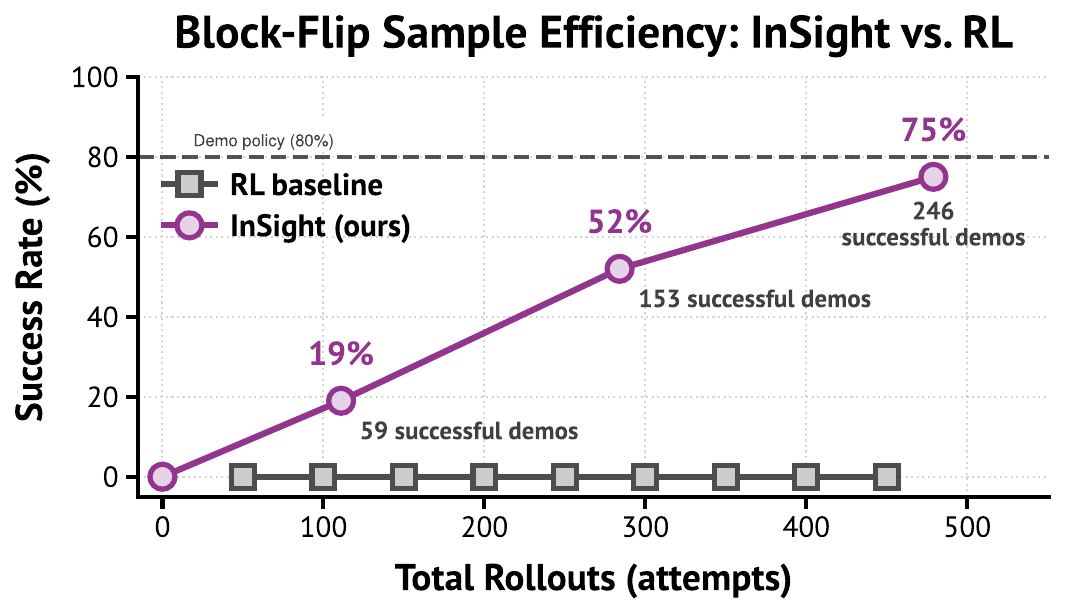}
\caption{\small{
\textbf{Block-flip skill acquisition with \method is more sample-efficient than RL.}
Starting from only pick-and-place demonstrations, \method autonomously acquires rotate-block examples. The retrained VLA reaches $75\%$ success with around 250 successfully acquired primitives.
The RL SAC \cite{haarnoja_soft_2018} baseline (given the same rollout budget) does not complete a flip (0\%); it learns to reach the block in 23\% of episodes and grasp it in 10\% of episodes, but never lifts and rotates to completion. The dotted line ($80\%$) is a VLA trained directly on human demonstrations of block flipping.}}
\label{fig:flip_iter}
\end{figure}

For this task, the robot is asked to flip a Lego block such that the peg faces up, given only 150 human teleoperated pick-and-place demonstrations of the block lying on its side. 
We automatically segment these demos into over 700 primitive episodes across seven primitive types. The block-flip task requires a \textit{rotate-block} primitive that is not present in pick-and-place demonstrations, which the VLM identifies as a primitive gap.

\textbf{In this setting, bootstrapping skill acquisition with primitives is more sample-efficient than our RL baseline, suggesting that \method can be a practical method for real-world skill acquisition.} Fig.~\ref{fig:flip_iter} shows that block-flip success improves as rotate-block primitive rollouts are added, reaching $75\%$ success after 246 acquired primitive rollouts, collected over 479 total attempts. On a comparable compute budget, an RL soft actor-critic (SAC) \cite{haarnoja_soft_2018} baseline never completes a flip. 

\subsection{Simulation: Drawer Closing from \textit{Only} Drawer-Opening Human Demos}

\begin{figure}[t]
\centering
\includegraphics[width=0.65\linewidth]{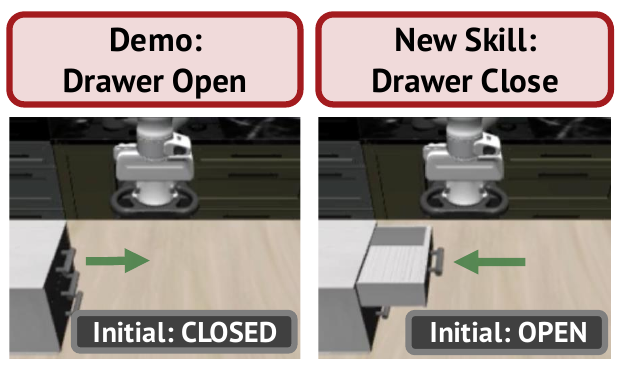}
\caption{\small{\textbf{Drawer closing.} A VLA is trained only on \emph{open-drawer} demos (left). Closing the drawer (right) requires a new \emph{push drawer closed} primitive executed from an open drawer, which is an initial state absent from the fine-tuning demonstrations. \method can use a VLM completion check to terminate the known approach primitive and trigger the new push drawer primitive.}}
\label{fig:drawer}
\end{figure}

The base policy is trained only on \textit{drawer-opening} demonstrations, and \method must acquire the missing \textit{push drawer closed} primitive starting from an open drawer (Fig.~\ref{fig:drawer}). 
We use 50 human teleoperated drawer-opening demonstrations, segmented into primitives: (1) \textit{move gripper to the drawer handle}, (2) \textit{close gripper}, and (3) \textit{pull the drawer open}.
This initial state for drawer closing is out-of-distribution (OOD) with respect to the fine-tuning demonstrations: the approach primitive was demonstrated only with the drawer initially closed, but the drawer closing task begins with the drawer already open. 
From an open drawer initial state, the approach primitive still reaches the handle, but the primitive may partially or fully close the drawer because the learned progress degrades outside the fine-tuning distribution. To handle this distribution shift, we use a VLM completion check that periodically queries whether the current primitive is complete.

\textbf{\method still acquires the missing primitive when a known primitive starts from an OOD initial state, using a VLM completion check to bridge imperfect primitive termination.}
Across 82 episodes, \method produces 70 successful push drawer closed primitives, where incorrect axis selection is the dominant primitive acquisition failure mode. Retraining a unified VLA on the original drawer-opening demonstrations with the new 70 push drawer closed primitives yields 100\% drawer closing success over 25 evaluation trials, while retaining the ability to open the drawer. The VLM completion check does not always align exactly with the semantic boundary between moving to the handle and pushing the drawer closed, but it provides a sufficient transition signal for reliable closed-loop drawer closing.

\begin{figure*}[t]
\centering
\includegraphics[width=1.0\textwidth]{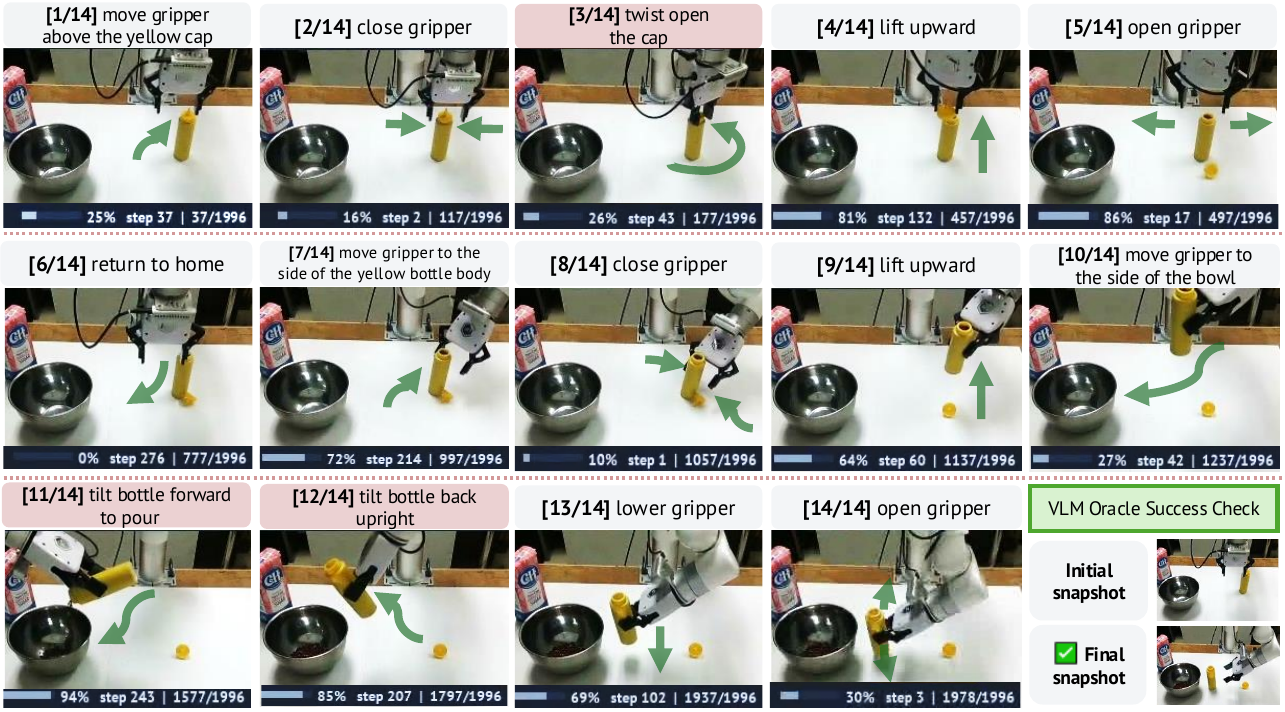}
\caption{\small{\textbf{Compositional twist-then-pour evaluation rollout.} \method chains 14 primitives from the separately acquired twist and pour skills, with no end-to-end demonstrations of the combined task and achieves $80\%$ end-to-end success. Shaded headers mark primitives acquired autonomously by \method and added back into the VLA's vocabulary; unshaded primitives are already known from the pick-and-place base demonstrations. The step/progress value shown in each panel is the learned \textit{per-primitive} progress channel, which resets at the start of each primitive rather than increasing monotonically across the full 14-primitive sequence. A VLM oracle verifies task success.}}
\label{fig:twist_pour}
\end{figure*}

\begin{figure*}[t]
\centering
\includegraphics[width=\linewidth]{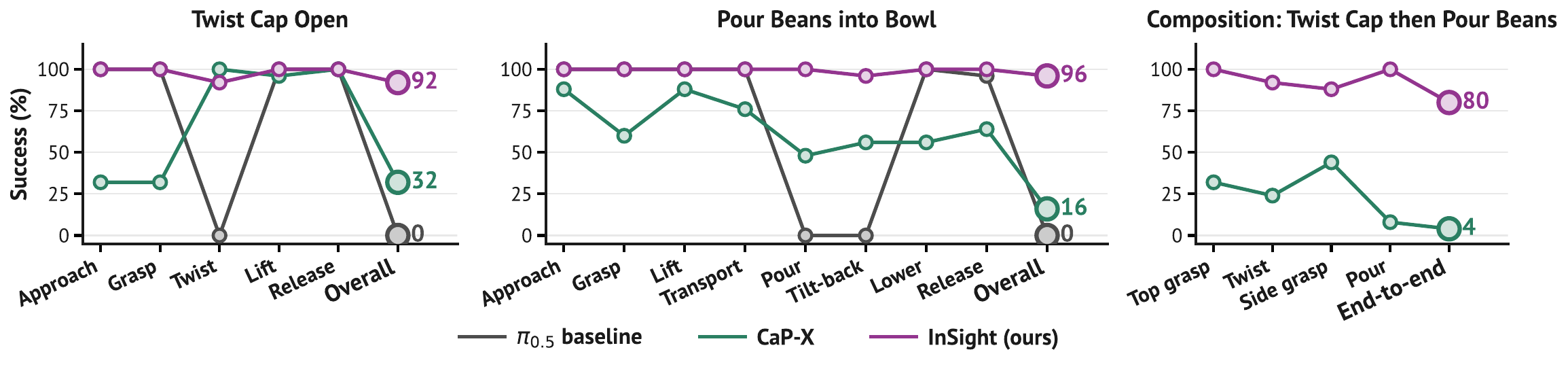}
\vspace{-6mm}
\caption{\small{\textbf{Per-primitive success rates for twisting and pouring} (25 trials per method). Each marker is the success rate of the labeled primitive across rollouts; \textit{Overall} / \textit{End-to-end} is full-task success.
The $\pi_{0.5}$ baseline is fine-tuned on 50 human pick-and-place demos;
\method additionally uses 20 successful acquired primitive episodes.
\method achieves $92\%$ twist and $96\%$ pour success, compared with $32\%$ and $16\%$ for CaP-X \cite{fu_cap-x_2026}. The twist and pour skills are composed without demonstrations of the combined task; \method achieves $80\%$ end-to-end success vs. $4\%$ for CaP-X.
}}
\label{fig:real_world_funnels}
\end{figure*}

\begin{figure}[t]
\centering
\includegraphics[width=\linewidth]{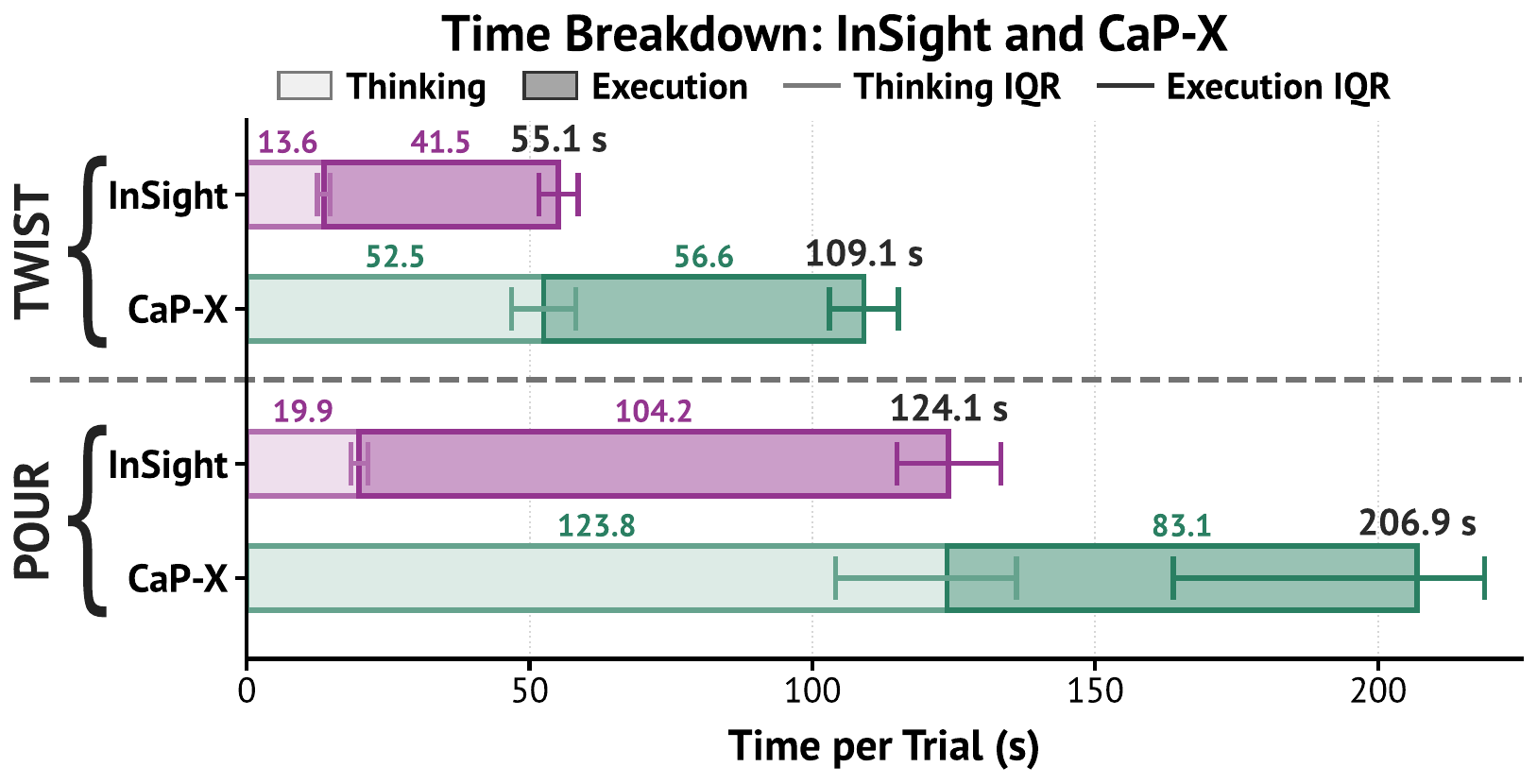}
\caption{\small{\textbf{Per-trial time decomposition} ($N=25$). After acquisition, \method spends substantially less wall-clock time per trial than CaP-X on both skills, since learned primitives are executed directly by the VLA rather than requiring repeated VLM-guided low-level control at test time. \textit{Thinking} reports VLM-call latency; \textit{Execution} reports robot motion.}}
\label{fig:efficiency_bars}
\end{figure}

\begin{figure}[t]
\centering
\includegraphics[width=0.8\linewidth]{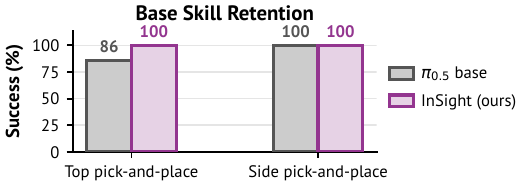}
\caption{\small{\textbf{Base skill retention} ($N{=}15$). After adding the acquired twist and pour primitives, the unified VLA retains its success on both original pick-and-place skills, showing no degradation on the tested base skills as the primitive vocabulary expands.}}
\label{fig:retention}
\end{figure}

\subsection{Real-World: Primitive Segmentation and Steerability} 
\label{sec:steerability}

We train a VLA on primitives automatically segmented from 50 human pick-and-place demonstrations (grasping the bottle from the top and side), and evaluate \method on three tasks that require primitives absent from this data: twisting open a bottle cap, pouring its contents into a bowl, and a long-horizon task that composes the twisting and pouring skills (as shown in Fig.~\ref{fig:twist_pour}). 

Before evaluating skill acquisition, we test Stage 1 (Sec.~\ref{sec:methods-stage1}): whether the automatic segmentation selects correct primitive boundaries and whether training on these segments exposes a primitive-steerable policy.

We first segment human demos into primitives to train a primitive-steerable VLA, and show that \textbf{\method's automatic primitive segmentation is comparable to manual labeling}. Across 20 demos from each of our two real-world pick-and-place human demo sets, automatic primitive segmentation boundaries differed from manual labeling by a mean of 6.04 frames ($0.3 \si{\second}$) and 5.71 frames ($0.29 \si{\second}$).

We then investigate primitive steerability by training on unsegmented vs. primitive-segmented pick-and-place human demos. Under primitive commands, the primitive-trained VLA correctly stops after each primitive in $16/16$ matched-start trials vs. $0/16$ for the unsegmented VLA, which instead continues the trajectory.


\subsection{Real-World: Skill Acquisition and Compositional Reuse}

\begin{table}[t]
\centering
\footnotesize
\setlength{\tabcolsep}{3pt}
\caption{\small{\textbf{Twist and pour primitive-acquisition cost.} 
Trials is the number of acquisition episodes to collect 20 successful primitive episodes. Robot-motion, VLM-call, and wall-clock are cumulative times (min), with per-trial means (seconds) in parentheses.}}
\label{tab:flywheel_efficiency}
\begin{tabular}{@{}lcccc@{}}
\toprule
Skill & Trials & Robot & VLM & Wall \\
\midrule
Twist & 23 & 23.8 (62\,s) & 8.4 (21.9\,s) & 39.7 (104\,s) \\
Pour  & 31 & 49.6 (96\,s) & 26.9 (52\,s)  & 85.3 (165\,s) \\
\bottomrule
\end{tabular}
\end{table}

Having established that training on primitive segmentations exposes a usable language interface for acquisition, we evaluate whether \method can leverage that interface to acquire new skills and compose the resulting primitives.
Given a new task, we collect rollouts using the steerable VLA for known primitives and a low-level controller parameterized by VLM-specified single-axis motions for new primitives. \method distills new primitives into the VLA for skill acquisition, with the VLM oracle selecting which acquisition rollouts are added to training.

\textbf{Each new skill is efficient to acquire}: Table~\ref{tab:flywheel_efficiency} shows 20 successful primitive episodes acquired in 23 trials for twisting and 31 for pouring. The primitive-steerable VLA is integral to the acquisition loop and downstream composition. During acquisition, the VLM-parameterized controller completes the full task in $87\%$ and $65\%$ of trials. After distillation, the retrained VLA executes the acquired skills with $92\%$ and $96\%$ success, respectively, in closed loop and without per-primitive VLM latency (Fig.~\ref{fig:real_world_funnels}). 


\textbf{Per-primitive reliability leads to high end-to-end success.} Fig.~\ref{fig:real_world_funnels} reports per-primitive success rates for the real-world twist, pour, and composition tasks. We compare against \mbox{CaP-X}~\cite{fu_cap-x_2026}, an instance of the broader class of methods that use a VLM to compose motions for zero-shot execution of new tasks at test time without expanding the learned policy, as well as $\pi_{0.5}$, which is fine-tuned only on human demonstrations and receives no new primitives from \method. 

\method maintains high success at every primitive, so per-primitive reliability compounds into 92\% and 96\% end-to-end success on twist and pour, compared to 32\% and 16\% for CaP-X, which uses the same VLM, task prompts, and setup. The $\pi_{0.5}$ baseline, fine-tuned on the same base demonstrations with no acquired primitives, fails at the twist and pour steps, confirming that these primitives are not reliably accessible from the base policy under these prompts. The gap remains large on the longer-horizon composition, where \method reaches 80\% end-to-end success while CaP-X drops to 4\%.

\textbf{\method improves on execution efficiency compared to the CaP-X baseline that performs each skill zero-shot}. 
Fig.~\ref{fig:efficiency_bars} breaks down per-trial timing into VLM-call latency and robot execution for \method and CaP-X. \method is more efficient during evaluation due to the acquired primitives being distilled into the closed-loop VLA.

\textbf{Primitives can be composed into new long-horizon skills for continual learning}.
As shown in Fig.~\ref{fig:twist_pour}, the robot must (1) grasp the bottle from the top, (2) twist to open the cap, (3) re-grasp the bottle from the side, and (4) pour beans into a bowl. This task tests sequential composition of separately acquired skills: \method must chain primitives from two different acquired skills, without having end-to-end demonstrations of the combined task. \method composes 14 primitives into a successful rollout, reaching 80\% end-to-end success compared to 4\% for CaP-X.



\textbf{\method preserves existing skills as new primitives are added.} As shown in Fig.~\ref{fig:retention}, even after adding the newly acquired twist and pour primitives, the unified VLA retains the original top- and side-pick-and-place skills.

\begin{figure}[b]
\centering
\includegraphics[width=0.95\linewidth]{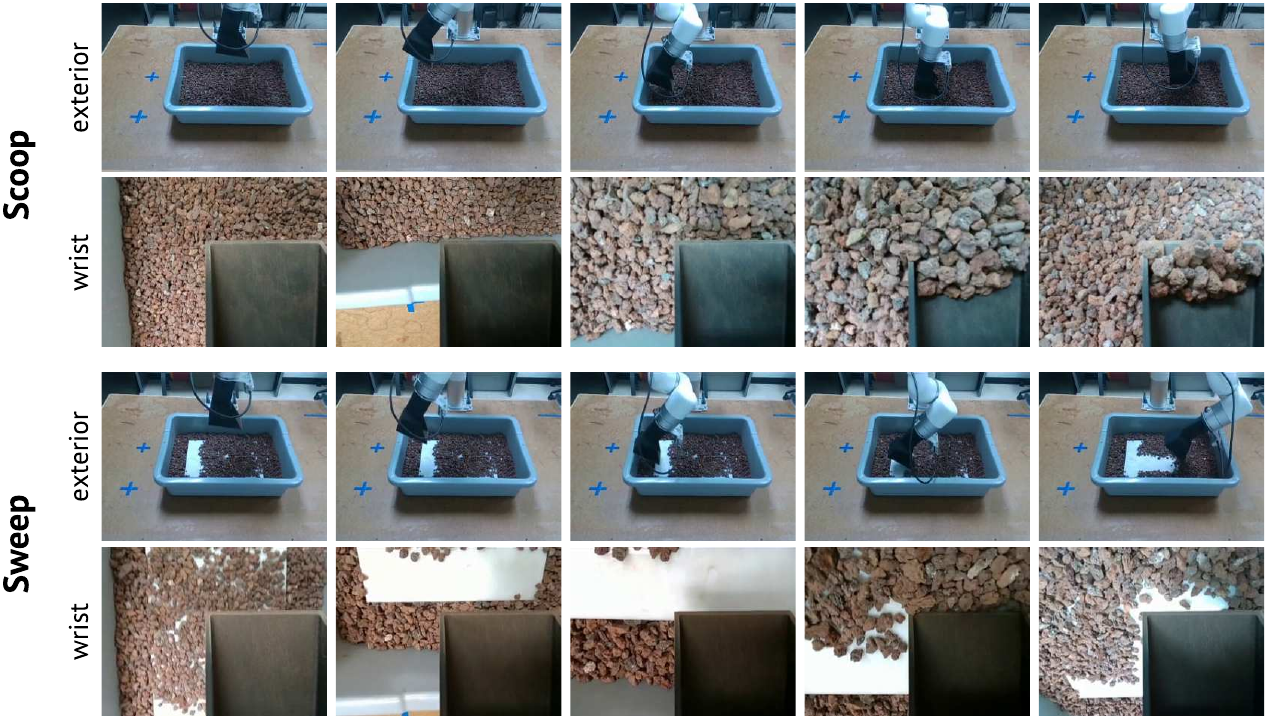}
\caption{\small{\textbf{Sweeping from only scooping human demonstrations.} Exterior and wrist views of the demonstrated scooping skill (top) and the sweeping skill acquired through \method (bottom). Since both scooping and sweeping require the gripper to be lowered to the rocks, \method acquires sweeping by adding a lateral-push primitive to the scooping primitives.}}
\label{fig:scoop_sweep}
\end{figure}

\subsection{Real-World: Sweeping from Scooping Human Demos}
\label{sec:sweeping_from_scooping}

\textbf{\method extends beyond grasping to contact-rich, non-prehensile motions.} This final task returns to our motivating scenario: a robot on Mars trained only to scoop rocks must sweep debris off its solar panels, with no sweeping demonstrations available. We evaluate whether \method can acquire a \textit{sweeping} primitive from demonstrations of scooping. Scooping and sweeping share the same approach and lowering primitives and differ only in the final motion after contact. While scooping requires the gripper to descend below the rocks, sweeping requires a lateral push across the surface, as shown in Fig.~\ref{fig:scoop_sweep}. The VLM flags this lateral push as the missing primitive. \method acquires the sweeping primitive and succeeds in all $5/5$ evaluation trials.

\section{Conclusion, Limitations, and Future Work}


We present \method, a method for autonomous skill acquisition in VLAs through VLM-guided primitive gap discovery and execution. By training on autonomously segmented primitives, identifying primitive gaps via VLM reasoning, and generating training data through VLM-guided low-level control, \method enables robots to acquire new skills without additional human demonstrations. 

Our results suggest several directions for future work. First, skill gap execution is restricted to single-axis motions, limiting the complexity of acquirable primitives. Future work would acquire richer primitives using VLM-generated waypoints, trajectory optimization, or online RL. Second, \method filters for successful rollouts; incorporating failure analysis and VLM feedback would allow for more sample-efficient skill acquisition \cite{zhai_vision-language-action-critic_2025}. In addition, manual environment resets are still necessary in this work. This cost can be mitigated by exploring rollouts with real-to-sim-to-real pipelines ~\cite{torne2024rialto} or a learned world model~\cite{ liu2026clap} that can test candidate rollouts before real-world execution, which can be especially crucial in safety-critical settings. Lastly, \method can be extended to embodiments with higher degrees of freedom, such as mobile manipulators or humanoids.

\section*{ACKNOWLEDGMENTS}

M.\ Wang is supported by the NASA NSTGRO Fellowship. S.\ Tian was supported by NSF GRFP Grant No. DGE-1656518.


\raggedbottom


\bibliographystyle{IEEEtran}
\bibliography{references}

\ifincludeappendix


\fvset{
    breaklines=true,
    breakanywhere=true,
    fontsize=\redboxfont,
    breaksymbolleft={},
    breaksymbolright={}
}

\useRomanappendicesfalse
\appendices
\label{appendix:xx}

\section{Implementation Details}
\label{app:implementation_details}


We use the $\pi_{0.5}$ VLA~\cite{intelligence2025pi05visionlanguageactionmodelopenworld} in our experiments, although \method is agnostic to the underlying VLA. We fine-tune with LoRA \cite{hu2021loralowrankadaptationlarge} (Gemma-2B backbone and Gemma-300M action expert with other weights frozen), with each segmented primitive as a separate training episode conditioned on its primitive label as the language prompt. The policy takes two $224\times224$ RGB views (scene and wrist) and the end-effector pose and gripper state, and outputs end-effector deltas, an absolute gripper command, and a learned progress channel $\in[0,1)$ supervised with the normalized timestep within each primitive segment. A primitive terminates when the progress channel exceeds a threshold (typically $0.95$), when end-effector motion falls below an auto-advance threshold, or (for out-of-distribution ``move to'' primitives) when a VLM completion check fires. 



\section{Scooping Implement Design \\and Experimental Setup}
\label{app:scooper_design}

The scooping implement used in the scooping to sweeping demonstration was inspired by the 2007 NASA Mars Phoenix Lander Robotic Arm and Icy Soil Acquisition Device~\cite{bonitz2008nasa}.
This sought to enable both regolith penetration (scooping), as well as rasping and chopping (different forms of sweeping), and was intended to yield information on Martian dry and icy soil properties such as angle of repose, cohesion, bearing strength, and grain size distribution.
The sweeping behavior was then motivated by the NASA Mars InSight lander's attempts to clear sand from its solar panels in 2021 when it was losing system power due to dust storms~\cite{nasa_insight_power_boost_2021}.

For the experimental setup, bulk particulates are placed into a container in order to train the initial scooping behavior.
A thin plastic sheet is then embedded at a shallow depth beneath the surface to emulate a submerged solar panel.
The scooping primitives of lowering and entering the rocks are then identified as missing a lateral pushing primitive which would be needed to complete the sweeping motion.

\section{VLM Demonstration Segmentation, Plan, Primitive Gap Proposal, and Oracle Check}


\method queries a vision-language model (Gemini 3 Flash~\cite{gemini3team2025}) in four roles, each constrained to return strict JSON: (i) offline \textbf{segmentation} of human demonstrations into primitive-labeled trajectories (\ref{app:demonstration_segmentation}); (ii) \textbf{planning}, which decomposes a new task into a primitive sequence and flags primitive gaps (\ref{app:task_planning}); (iii) \textbf{primitive-gap proposal}, which maps each gap to a single-axis motion the low-level controller executes (\ref{app:primitive_gap_proposal}); and (iv) \textbf{oracle} checks that verify primitive- and task-completion from images (\ref{app:oracle_and_completion_check}). The system prompts are reproduced verbatim below (\texttt{\{...\}} are runtime-filled fields).

\subsection{Demonstration Segmentation}
\label{app:demonstration_segmentation}
Demonstrations are segmented offline in three stages. First, the VLM decomposes the task instruction into an ordered primitive sequence. Second, the subsampled video is passed frame-by-frame and each frame is assigned to a plan primitive, cross-checking the image against a per-frame end-effector motion caption that reports the dominant translation/rotation axis, then returns the boundary frames between consecutive primitives. Each frame caption has the form:
\begin{tcolorbox}[stanfordstyle, breakable, title={Per-frame motion caption (segmentation input)}]
\begin{Verbatim}
Frame 120 | EE x=+0.213 y=-0.041 z=+0.118 rx=+3.04 ry=-0.02 rz=+1.57  | d dx=+0.004 dy=-0.001 dz=-0.012 |dxy|=0.004 |dz|=0.012  |drxy|=0.01 |drz|=0.00 dom=-z
\end{Verbatim}
\end{tcolorbox}
Third, each boundary is refined by a localized pass that reconciles the end-effector delta change-point with the earliest visually unambiguous frame. The result is a set of contiguous, primitive-labeled segments, each of which becomes one training episode (Appendix~\ref{app:implementation_details}).

\subsection{Task Planning and Primitive-Gap Flagging}
\label{app:task_planning}
Given a goal and the current primitive vocabulary, the planner returns the full primitive sequence, an execution note per step, and the subset of steps that are new (primitive gaps; the \texttt{skill\_gaps} field below). Each primitive gap is constrained to a single-axis motion, so a multi-motion task yields multiple gaps.

\begin{tcolorbox}[stanfordstyle, breakable, title={System prompt: PLAN\_TASK}]
\begin{Verbatim}
You are a robot task planner. Scene: {scene_context}

AVAILABLE PRIMITIVES (each is general-purpose and adapts to context): 
{primitives}

RULES:
1. Break the goal into fine-grained steps. Use existing primitives for every sub-step they cover -- a skill gap should only be the novel part, not a bundle of existing + novel actions.
2. Only create a skill gap when the desired outcome is fundamentally different from what any existing primitive produces. If an existing primitive could achieve the same result (even if executed differently), use it and put execution details in step_notes instead.
3. Every step goes in primitive_sequence -- including new ones.
4. New primitives also go in skill_gaps (must appear in BOTH lists).
5. Name new primitives by their desired EFFECT, not the robot motion.
6. For each step, add a note on execution (approach, grasp, how it enables
   the next step).
7. After the final step, the runtime returns the gripper to a safe home pose, so the gripper does not need to be cleared from the workspace by a final step in the plan. Each step should make a distinguishable contribution to the goal -- avoid adding a final step whose only effect is repositioning the gripper.
8. Each skill gap is one single-axis motion (one translation OR one rotation along one axis, in one direction). If the goal involves multiple distinct motions, create a separate skill gap for each.

Example 1 -- pick and place (all existing, no skill gaps):
  primitive_sequence: ["move gripper to the red lego block", "close gripper", "lift upward", "move gripper to target", "lower gripper", "open gripper"]
  skill_gaps: []

Example 2 -- inserting an object (one new skill gap):
  primitive_sequence: ["move gripper to object", "close gripper", "lift upward", "move gripper to target", "insert object into slot", "open gripper"]
  skill_gaps: ["insert object into slot"]

Respond with ONLY valid JSON:
{"primitive_sequence": ["step1", "step2", ...],
  "step_notes": ["execution note for each step"],
  "skill_gaps": ["new primitives not in available list"],
  "reasoning": "brief explanation",
  "confidence": 0.0-1.0,
  "requires_new_primitive": true or false}
\end{Verbatim}
\end{tcolorbox}

\par\smallskip\noindent\textbf{Example output} (pour data collection; the planner detects two rotational gaps):
\begin{tcolorbox}[stanfordstyle, breakable, title={Example output: PLAN\_TASK}]
\begin{Verbatim}
{"primitive_sequence": ["move gripper to the side of the bottle",
   "close gripper", "lift upward", "move gripper to the side of the bowl",
   "pitch bottle forward to pour", "tilt bottle back upright",
   "lower gripper", "open gripper"],
 "skill_gaps": ["pitch bottle forward to pour", "tilt bottle back upright"],
 "reasoning": "The existing primitives cover linear movement, grasping, and
   lifting. However, the pouring action requires a rotational movement (pitch) not in the available list. Per Rule 8, each single-axis motion in one direction is its own skill gap, so the forward pitch and the backward tilt to upright are two separate gaps.",
 "confidence": 1.0,
 "requires_new_primitive": true}
\end{Verbatim}
\end{tcolorbox}
A full executed plan (the twist-then-pour composition) is shown in Appendix~\ref{app:vlm_plans}.

\subsection{Primitive-Gap Proposal}
\label{app:primitive_gap_proposal}
For each flagged gap, the VLM sees the exterior and wrist images and returns a single-axis motion: an \texttt{axis} (\texttt{dx,dy,dz} base-frame translation or \texttt{drx,dry,drz} gripper-local rotation), a signed magnitude in meters or degrees, and an \texttt{already\_complete} flag. The controller then drives the arm along that one axis until the completion check (below) fires.
\begin{tcolorbox}[stanfordstyle, breakable, title={System prompt: PREANALYZE}]
\begin{Verbatim}
Determine the single-axis MOTION (translation OR rotation) needed to achieve the GOAL on a real xArm. The controller drives the arm along whichever single axis you pick.

IMAGES:
- IMAGE 1 (exterior overview): use for TRANSLATION reasoning.
- IMAGE 2 (wrist down-view, moves with the gripper): use for ROTATION axis selection.

TRANSLATION (dx, dy, dz) is in the BASE frame: +X = forward (into workspace), +Y = left, +Z = up.
ROTATION (drx, dry, drz) is in the GRIPPER's LOCAL frame (rotate WITH gripper):
  drz: Axial twist around the gripper's local Z axis (the camera's line of sight). Spins the held object in place around its own centerline like a screwdriver. It CANNOT pivot, tilt, or invert the angle of an object.
  dry: Pitch rotation around the gripper's local Y axis. Tilts/nods the gripper body forward or backward along its opening/closing path.
  drx: Roll rotation around the gripper's local X axis. Tilts the gripper body laterally sideways.

To pick the rotation axis: Do NOT pick the axis from the global room frame, camera frame, or verbs like "sideways/forward" -- look at IMAGE 2, since robot base frame coordinates differ from gripper coordinates.
- If the object needs to pivot/tilt forward/backward relative to the gripper body, pick dry.
- If the object needs to tilt laterally sideways relative to the gripper body, pick drx.
- If the object needs to twist/spin in place along its own centerline without changing its physical tilt angle, pick drz.

GEOMETRIC CONSTRAINT WARNING:
1. Never select drz for any motion that requires an object to tip over, invert, or pivot its top towards a target; drz only spins the object on its own axis.
2. The wrist camera moves with the gripper; its local axes are independent of the global room frame. Never select an axis based on where a target appears to sit (left, right, up, down) in IMAGE 1. Map the required tilt strictly to the local structure of the gripper fingers in IMAGE 2.

BE AWARE: Depth and gripper biases may exist due to the close-up wrist view.

SIGNED MAGNITUDE:
- Translation: meters, typical 0.05-0.20 m.
- Rotation:    degrees, typical 30-180 deg.
- Sign: along the chosen axis.

If already complete, set already_complete=true.

Respond with ONLY valid JSON:
{"current_state": "...",
  "target_state": "...",
  "reasoning": "...",
  "axis": "dx" | "dy" | "dz" | "drx" | "dry" | "drz",
  "signed_magnitude_m": 0.0,
  "signed_magnitude_deg": 0.0,
  "already_complete": false}

Set signed_magnitude_m=0 for rotation; signed_magnitude_deg=0 for translation.
\end{Verbatim}
\end{tcolorbox}

\par\smallskip\noindent\textbf{Example outputs} (two resolved gaps from logged runs):
\begin{tcolorbox}[stanfordstyle, breakable, title={Example output: PREANALYZE --- ``unscrew bottle cap''}]
\begin{Verbatim}
{"current_state": "The gripper is securely holding the cap of the yellow bottle.",
 "target_state": "The cap is rotated 150 degrees counter-clockwise to detach it from the bottle threads.",
 "reasoning": "Unscrewing is an axial twist of the held cap about the gripper's line of sight (local Z), so drz; ~150 deg detaches it.",
 "axis": "drz",
 "signed_magnitude_m": 0.0,
 "signed_magnitude_deg": 150.0,
 "already_complete": false}
\end{Verbatim}
\end{tcolorbox}
\begin{tcolorbox}[stanfordstyle, breakable, title={Example output: PREANALYZE --- ``pitch bottle forward to pour''}]
\begin{Verbatim}
{"current_state": "The gripper is holding the yellow bottle in a side-grasp, positioned next to the metal bowl.",
 "target_state": "The yellow bottle is tilted over the bowl so that all its contents are poured out.",
 "reasoning": "To pour the contents into the bowl, the robot needs to rotate the gripper. Looking at IMAGE 2 (the wrist camera view), the bowl is positioned 'below' the bottle. A pitch rotation around the gripper's local Y-axis (dry) tilts the gripper body and the held bottle forward/downward into the bowl. A magnitude of 110 degrees ensures the bottle is tilted enough to empty all the beans. drz is excluded by the geometric constraints because the motion requires tipping the bottle over.",
 "axis": "dry",
 "signed_magnitude_m": 0.0,
 "signed_magnitude_deg": 110.0,
 "already_complete": false}
\end{Verbatim}
\end{tcolorbox}

\subsection{Oracle and Completion Checks}
\label{app:oracle_and_completion_check}
After a plan executes, an \emph{oracle} compares the initial and final scene images and accepts the trial only if the task is achieved; accepted trials become training demonstrations and the rest are discarded. 
\begin{tcolorbox}[stanfordstyle, breakable, title={System prompt: TASK\_COMPLETION (oracle)}]
\begin{Verbatim}
You are a vision system inspecting a robot manipulation scene from a fixed external camera. Compare two images (BEFORE and AFTER a robot action) and decide whether the task was completed.
\end{Verbatim}
\end{tcolorbox}
\noindent paired with the user message:
\begin{tcolorbox}[stanfordstyle, breakable, title={User message: oracle}]
\begin{Verbatim}
IMAGE 1 = BEFORE the action (gripper at home, scene in initial state).
IMAGE 2 = AFTER the action.

Task: {task}.

Question: Was the task completed?

Respond with valid JSON only:
{"completed": true | false, "reasoning": "<one sentence>"}
\end{Verbatim}
\end{tcolorbox}

\par\smallskip\noindent\textbf{Example oracle verdicts} (accepted twist-then-pour trials):
\begin{tcolorbox}[stanfordstyle, breakable, title={Example output: TASK\_COMPLETION}]
\begin{Verbatim}
{"completed": true, "reasoning": "The yellow bottle is uncapped, its contents have been poured into the bowl, and the bottle has been returned to the table with the gripper now open."}

{"completed": true, "reasoning": "The bottle has been successfully uncapped, moved from its initial position to the bowl, and released, and the gripper has returned to an open state at the home position."}
\end{Verbatim}
\end{tcolorbox}
\begin{tcolorbox}[stanfordstyle, breakable, title={System prompt: PRIMITIVE\_DONE (completion check)}]
\begin{Verbatim}
Determine if the robot primitive has been completed.

You receive two images:
- IMAGE 1 (exterior camera): side view across the table. PRIMARY signal for depth and height.
- IMAGE 2 (wrist camera): top-down from the gripper. Use ONLY for centering/identification.

CRITICAL: Top-down views (IMAGE 2) make objects appear close even when there is a large vertical gap. ALWAYS judge vertical proximity from IMAGE 1 (side view). If you can see ANY visible vertical gap between the gripper bottom and the target in IMAGE 1, the primitive is NOT done -- even if IMAGE 2 shows them overlapping.

For "move gripper to X" or "touch X":
- Done = gripper bottom is contacting X or within ~5mm of X in IMAGE 1.
- NOT done = visible vertical gap between gripper bottom and target in IMAGE 1.

For state-change primitives (open/close/push/pull/rotate/scoop/sweep/lift):
- Done = the target state is visibly achieved (drawer closed, object pushed away, rocks displaced, etc.) in IMAGE 1.

Reasoning must explicitly describe what you see in IMAGE 1, not just IMAGE 2.

Respond with ONLY JSON: {"done": true or false,
  "reasoning": "brief, references IMAGE 1"}
\end{Verbatim}
\end{tcolorbox}

\section{VLM-Generated Plans}
\label{app:vlm_plans}

The following is an example VLM-generated plan for the twist-then-pour compositional task. Note that all of the primitives are marked as \textit{known} because they have all been added to the VLA, including the primitives generated through \method (\textit{twist open the cap}, \textit{tilt bottle forward to pour}, \textit{tilt bottle back upright}).

\begin{tcolorbox}[
  stanfordstyle,
  breakable,
  title={Primitive Plan for Twist-then-Pour},
  label={box:twist_pour_plan}
]
\textbf{Task:} Open the yellow bottle cap, regrasp the bottle, and pour its contents into the bowl.

\vspace{0.4em}

\begin{enumerate}[leftmargin=2.2em,itemsep=0.35em,topsep=0.2em]
    \item \knownbadge\ \textbf{move gripper above the yellow bottle cap} --- Move the gripper into a top-down approach position centered over the yellow cap.
    \item \knownbadge\ \textbf{close gripper} --- Close the gripper to secure a firm grasp on the cap.
    \item \knownbadge\ \textbf{twist open the cap} --- Perform a 180-degree counterclockwise rotation to unscrew the cap from the bottle.
    \item \knownbadge\ \textbf{lift upward} --- Lift the cap vertically to ensure it is completely detached from the bottle threads.
    \item \knownbadge\ \textbf{open gripper} --- Open the gripper to drop the detached cap onto the workspace.
    \item \knownbadge\ \textbf{return to home} --- Execute the mandatory hardware reset to return the robot to its canonical home pose.
    \item \knownbadge\ \textbf{move gripper to the side of the yellow bottle body} --- Move the gripper to a side-approach position relative to the bottle body.
    \item \knownbadge\ \textbf{close gripper} --- Close the gripper to perform a side grasp on the now-uncapped bottle.
    \item \knownbadge\ \textbf{lift upward} --- Lift the bottle upward to clear the table for movement.
    \item \knownbadge\ \textbf{move gripper to the side of the bowl} --- Transport the bottle to the side of the bowl in preparation for pouring.
    \item \knownbadge\ \textbf{tilt bottle forward to pour} --- Tilt the bottle forward over the bowl to empty its contents.
    \item \knownbadge\ \textbf{tilt bottle back upright} --- Rotate the bottle back to a vertical, upright orientation.
    \item \knownbadge\ \textbf{lower gripper} --- Lower the bottle back down to the table surface.
    \item \knownbadge\ \textbf{open gripper} --- Open the gripper to release the bottle.
\end{enumerate}
\end{tcolorbox}







\fi 

\end{document}